\documentclass{article}
\usepackage{arxiv}
\usepackage[T1]{fontenc}    
\usepackage{hyperref}       
\usepackage{url}            
\usepackage{booktabs}       
\usepackage{amsfonts}       
\usepackage{nicefrac}       
\usepackage{microtype}      
\usepackage{lipsum}		
\usepackage{graphicx}
\usepackage{natbib}
\usepackage{booktabs}
\usepackage{subfig}
\usepackage{doi}

\title{Malayalam Sign Language Identification using Finetuned YOLOv8 and Computer Vision Techniques}

\author{K. Abhinand\\
    NLP for Social Good Lab\\
	School of Digital Sciences\\
	Kerala University of Digital Sciences-\\Innovation and Technology\\
	Thiruvananthapuram, India \\
	\texttt{abhinand.ds22@duk.ac.in} \\
	\And
	Abhiram B. Nair \\
    NLP for Social Good Lab\\
	School of Digital Sciences\\
	Kerala University of Digital Sciences-\\Innovation         and Technology\\
	Thiruvananthapuram, India \\
	\texttt{abhiram.ds23@duk.ac.in} \\
	\And
	C. Dhananjay \\
    NLP for Social Good Lab\\
	School of Digital Sciences\\
	Kerala University of Digital Sciences-\\Innovation         and Technology\\
	Thiruvananthapuram, India \\
	\texttt{dhananjay.ds22@duk.ac.in} \\
        \And
	Hanan Hamza \\
    NLP for Social Good Lab\\
	School of Digital Sciences\\
	Kerala University of Digital Sciences-\\Innovation         and Technology\\
	Thiruvananthapuram, India \\
	\texttt{hanan.ds22@duk.ac.in} \\
    \And
	J. Mohammed Fawaz \\
    NLP for Social Good Lab\\
	School of Digital Sciences\\
	Kerala University of Digital Sciences-\\Innovation         and Technology\\
	Thiruvananthapuram, India \\
	\texttt{mohammedfwz.ds22@duk.ac.in} \\
    \And
	K. Rahma Fahim \\
    NLP for Social Good Lab\\
	School of Digital Sciences\\
	Kerala University of Digital Sciences-\\Innovation         and Technology\\
	Thiruvananthapuram, India \\
	\texttt{rahma.ds22@duk.ac.in} \\
 \And
	V. S. Anoop \\
    NLP for Social Good Lab\\
	School of Digital Sciences\\
	Kerala University of Digital Sciences-\\Innovation         and Technology\\
	Thiruvananthapuram, India \\
	\texttt{anoop.vs@duk.ac.in} \\
}

\date{}

\renewcommand{\shorttitle}{Malayalam Sign Language Identification using Computer Vision Techniques}
\begin{document}
\maketitle

\begin{abstract}
Technological advancements and innovations are advancing our daily life in all the ways possible but there is a larger section of society who are deprived of accessing the benefits due to their physical inabilities. To reap the real benefits and make it accessible to society at large, these specially talented and specially gifted people should also use such innovations without any hurdles. Many applications developed these days address these challenges, but localized communities and other constrained linguistic groups may find it difficult to use them. Malayalam, a Dravidian language spoken in the Indian state of Kerala is one of the twenty-two scheduled languages in India. Recent years witnessed a surge in the development of systems and tools in Malayalam, addressing the needs of Kerala, but many of them are not empathetically designed to cater to the needs of hearing-impaired people. One of the major challenges is the limited or no availability of sign language data for the Malayalam language and sufficient efforts are not made in this direction. In this connection, this paper proposes an approach for sign language identification for the Malayalam language using advanced deep learning and computer vision techniques. We start by developing a labeled dataset for Malayalam letters and for the identification we use advanced deep learning techniques such as YOLOv8 and computer vision. Experimental results show that the identification accuracy is comparable to other sign language identification systems and other researchers in sign language identification can use the model as a baseline to develop advanced models.
\end{abstract}
\section{Introduction}
In the contemporary landscape of technological innovation, strides in various domains have revolutionized human existence, from education\cite{abulibdeh2024navigating} and healthcare\cite{younis2024systematic} to entertainment\cite{takale2024advancements} and communication\cite{gholami2024rise}. Despite these transformative advancements, certain communities continue to face barriers hindering their access to basic privileges. Communication is a vital aspect of human interaction, and for individuals with hearing or speaking disabilities, sign language serves as a crucial means of expressing thoughts and ideas\cite{almufareh2024conceptual}. The sign language users in Kerala face the challenge of relying on broader sign languages such as Indian Sign Language (ISL), American Sign Language (ASL), or other regional sign languages to communicate effectively until very recently\cite{renjith2024sign}\cite{priya2024developing}. The absence of a dedicated sign language tailored to the linguistic and cultural nuances of Kerala posed a barrier to seamless communication for the deaf community in the region. The National Institute of Speech and Hearing (NISH)\footnote{https://www.nish.ac.in/}, an institute devoted to the education and rehabilitation of individuals with speech-language and hearing impairments, located in Thiruvananthapuram, introduced Malayalam Sign Language (MSL) in September 2021, marking a great advancement in inclusivity and accessibility. This project strives to empower the hearing and hearing-impaired community in Kerala through the deployment of a sophisticated sign language identification model. The primary objective of this study is to develop a robust artificial intelligence model capable of recognizing static hand gestures specific to Malayalam Sign Language from real-time videos. By utilizing the advanced capabilities of Computer Vision, the model generates corresponding captions, facilitating enhanced communication for individuals within the hearing and speech-impaired community.
\begin{figure}[!h]
    \centering
    \includegraphics[width=\textwidth]{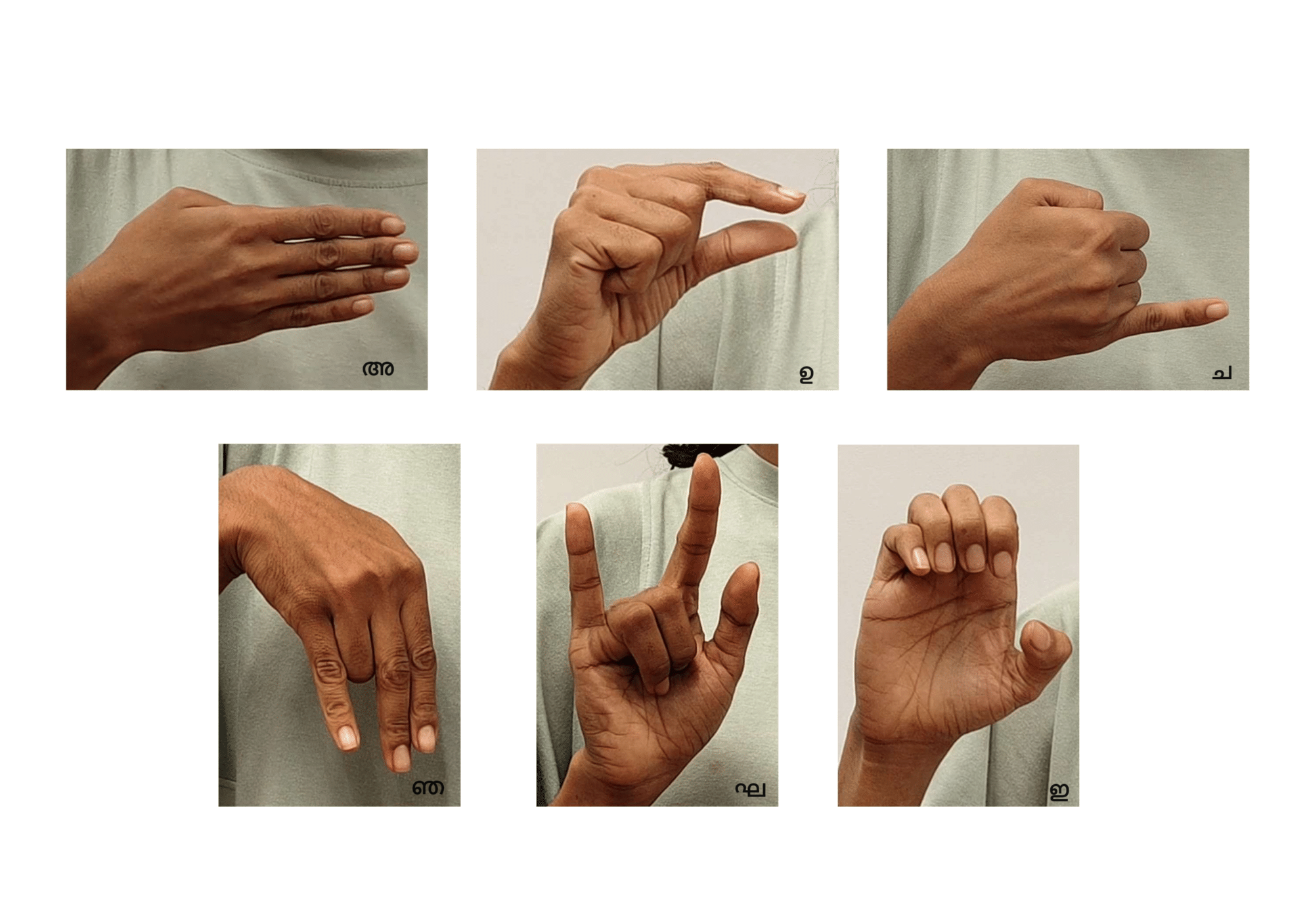}
    \caption{Examples of 6 Malayalam Sign Language (MSL) characters}
    \label{fig:nish-alphabet}   
\end{figure}
Malayalam Sign Language was introduced in Kerala in September 2021, providing a welcome improvement for individuals with hearing or speaking disabilities. \cite{issac2023deep} elaborates on this and introduces a real-time Sign Language Recognition system using transfer learning with TensorFlow. It uses a webcam to detect and interpret Malayalam characters from MSL, aiming to improve communication for individuals with speech impairments. \cite{praneel2023malayalam} discusses the same, presenting a character recognition system using a modified Inception V4 model to accurately identify nine Malayalam characters in MSL communication. The performance of this system surpasses other state-of-the-art methods, achieving a Top-1 error rate of 17.7\% and a Top-5 error rate of 3.8\%.
\cite{salim2023sign} also uses transfer learning to interpret characters form MSL. Their proposed method employs ResNet50 to classify static sign language alphabet images with a training accuracy of 97.62\% and validation accuracy of 92.35\%.

YOLO is a state-of-the-art object detection algorithm popular for its real-time processing capability, accurately identifying and localizing multiple objects within images or videos in a single pass \cite{redmon2016look}. YOLOv8 is the latest in this series of models, with improved accuracy, speed, and versatility compared to earlier versions. \cite{tyagi2023american} explores the application of these to detect American Sign Language (ASL) gestures. It compares different YOLO versions, emphasizing YOLOv8’s superior precision and mean Average Precision (mAP). Trained and tested on the American Sign Language letters dataset, the model achieves 95\% precision, 97\% recall, and 96\% mAP@0.5. \cite{jiaslr} proposes an enhanced SLR-YOLO network for efficient sign language recognition. Addressing challenges of traditional methods, the model builds upon YOLOv8, incorporating modifications such as replacing the SPPF module is replaced with RFB module for enhanced feature extraction and introducing BiFPN and Ghost module for improved feature fusion and reduced network weight. Data generalization is enhanced throughout the Cutout method during training, resulting in improved accuracy on validation sets: 90.6\% for American Sign Language letters and 98.5\% for Bengali Sign Language Alphabet, surpassing the performance of the original YOLOv8. The improved model achieves a 1.3\% accuracy increase, an 11.31\% reduction in parameters, and an 11.58\% reduction in FLOPs.\cite{vidhyasagar2023video} focuses on the rapid growth in sign language recognition research. Using the Roboflow dataset and Transfer Learning with the YOLOv8, the study proposes a real-time transcription system for ASL signs (A to Z) during live meetings or video conferences. The model demonstrates effective communication by extracting essential components from input video frames and classifying signs based on neural network comparisons. This paper presents a Malayalam sign language identification model that identifies Malayalam Sign Language characters from videos and displays them for easy interpretation. The core contributions of this paper are summarized as follows:
\begin{itemize}
    \item Discusses the relevance of sign language identification for hearing-impaired people.
    \item Reviews some state-of-the-art approaches for sign language identification.
    \item Proposes an approach for sign language identification for Malayalam language using YOLOv8 and computer vision techniques. 
    \item Experimentally verifies the proposed system to showcase the usefulness in developing inclusive artificial intelligence applications.
\end{itemize}
The structure of the paper is as follows: Section 2 presents the materials and methods, Section 3 presents the proposed approach, Section 3 outlines the experiment and in Section 5 the results and discussions are presented. In section 6, the authors conclude the paper and presents some dimensions for future research.
\section{Materials and Methods}
\subsection{Roboflow}
Roboflow, available at \url{https://roboflow.com/}, is a computer vision platform that enables users to build models efficiently by offering improved methods for data collection, preprocessing, and model training. It is an extremely useful tool for computer vision researchers and developers want to streamline and improve the intricate process of preprocessing and managing datasets. Managing and preparing datasets for machine learning models is a challenging task, which is the fundamental purpose of Roboflow. Integrating it easily with well-known computer vision frameworks like TensorFlow, PyTorch, and YOLO is one of its standout characteristics. By giving users the option to work with a variety of model architectures and frameworks, this integration promotes creativity and adaptation in the quickly developing field of computer vision. Data augmentation and preprocessing are critical processes in improving dataset quality and diversity, which directly affect the performance of machine learning models. Roboflow excels in these domains, providing a diverse collection of augmentation strategies for enriching datasets. The platform's user-friendly interface enables users to easily apply transformations, resize photos, and perform other preprocessing procedures, saving critical time in the data preparation workflow. Roboflow is a shining example of efficiency in the complicated field of computer vision. It is a top option for academics and developers due to its integration capabilities, dataset management tools, and dedication to data quality through augmentation and pretreatment. Roboflow continues to be a dependable ally, enabling users to confidently and easily negotiate the complexities of dataset preparation as technology develops and machine learning applications become more complex.

\subsection{Ultralytics}: Ultralytics, available at \url{https://www.ultralytics.com/}, is a versatile toolkit for computer vision, compatible with popular frameworks like PyTorch and TensorFlow. It simplifies model development and training with a user-friendly interface and supports various model architectures for diverse project requirements. Ultralytics YOLOv8 (\url{https://www.ultralytics.com/yolo}), an advanced object detection model, is a key component within the Ultralytics toolkit, renowned for its real-time efficiency and high accuracy. Seamlessly integrating with the Ultralytics ecosystem, YOLOv8 stands out for its versatility, adapting to diverse object detection tasks in domains such as autonomous vehicles, surveillance, and industrial automation. The model features an enhanced architecture, incorporating elements like the CSPDarknet53 backbone and PANet for improved feature extraction, contributing to heightened accuracy and robustness. Benefiting from the optimization techniques within the Ultralytics toolkit, YOLOv8 employs mixed-precision training and distributed training, facilitating faster convergence during training and efficient resource utilization, even for extensive datasets. Its success extends to real-world applications, demonstrating efficacy in detecting objects across various domains. Maintaining a user-friendly interface, YOLOv8 aligns with Ultralytics' commitment to simplicity without compromising on the model's depth for complex computer vision tasks. The model's active engagement with the Ultralytics community ensures continuous improvement, reflecting its resilience and user-friendly nature. In conclusion, Ultralytics YOLOv8 represents a significant stride in object detection models, offering researchers and developers a powerful tool to address diverse and challenging real-world scenarios. 
\subsection{YOLOv8}
YOLOv8 is a real-time object detection and image segmentation model based on cutting-edge advances in deep learning and computer vision, providing exceptional speed and accuracy. Its simplified architecture makes it appropriate for a wide range of applications and easily scalable to a variety of hardware platforms, from edge devices to cloud APIs. A breakdown of its key components is given below:
\begin{figure}[h]
    \centering
    \includegraphics[width=\textwidth]{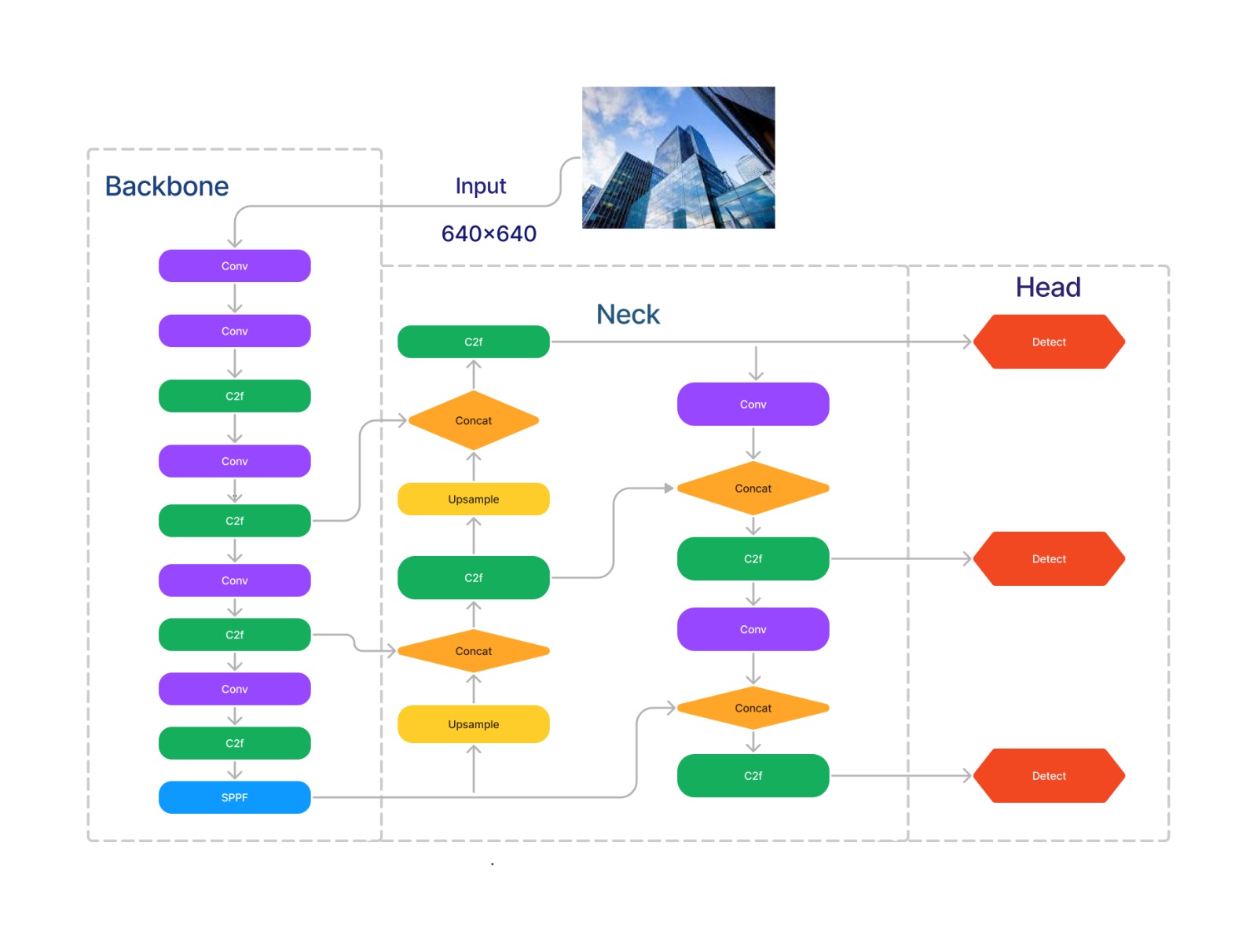}
    \caption{A diagrammatic representation of YOLOv8 architecture} 
    \label{fig:yoloarc}   
\end{figure}

\begin{itemize}
    \item Backbone
    \begin{itemize}    
        \item Modified CSPDarknet53: The core of YOLOv8 lies in its backbone, a modified version of the CSPDarknet53 architecture. This network features 53 convolutional layers, carefully chosen and optimized for both performance and memory footprint.
        \item Cross-Stage Partial Connections (CSP): This innovative technique reduces redundant computations while maintaining information flow through the network.
    \end{itemize}
    \item Neck
    \begin{itemize}
        \item SPPF and New CSP-PAN: This section connects the backbone and the head, further enriching spatial features and enhancing feature propagation.
    \end{itemize}
    \item Head
    \begin{itemize} 
        \item Predicting Bounding Boxes and Class Probabilities: Additional convolutional layers followed by fully connected layers process information from the neck and predict bounding boxes, objectness scores, and class probabilities for detected objects.
    \end{itemize}
\end{itemize}
YOLOv8 processes objectness, classification, and regression tasks independently using an anchor-free model with a decoupled head. This design allows each branch to focus on its task and improves the model's overall accuracy. The sigmoid function was utilized as the activation function for the objectness score in the YOLOv8 output layer, expressing the probability that the bounding box includes an object. It use the softmax function to represent the probabilities of items belonging to each conceivable class. YOLOv8 uses CIoU [68] and DFL [108] loss functions for bounding box loss and binary cross-entropy for classification loss. These losses have enhanced object detection performance, especially with tiny objects. Instead of the typical YOLO neck architecture, the backbone is a CSPDarknet53 feature extractor followed by a C2f module. Following the C2f module are two segmentation heads that learn to anticipate the semantic segmentation masks for the input image. The model has similar detection heads to YOLOv8, consisting of five detection modules and a prediction layer. The YOLOv8-Seg model has achieved cutting-edge performance on a variety of object recognition and semantic segmentation benchmarks while remaining fast and efficient.

\subsection{Dataset}
\label{ref_dataset}
The dataset comprises images extracted from video frames, capturing static signs of Malayalam sign language across 20 distinct signs. These videos, converted into frames at a rate of 60 frames per second, yielded 100 photos for each sign. In the preprocessing phase, the images were resized, 5\% noise was introduced and a rotation range of ten degree was applied during data augmentation. Post-preprocessing, the dataset encompassed 5900 data points, including both positive and negative rotations, enhancing the dataset's diversity and robustness for effective model training. Using Roboflow, we conducted the annotation process, formatting the data in YOLO.
\begin{figure}[h]
    \centering
    \includegraphics[width=\textwidth]{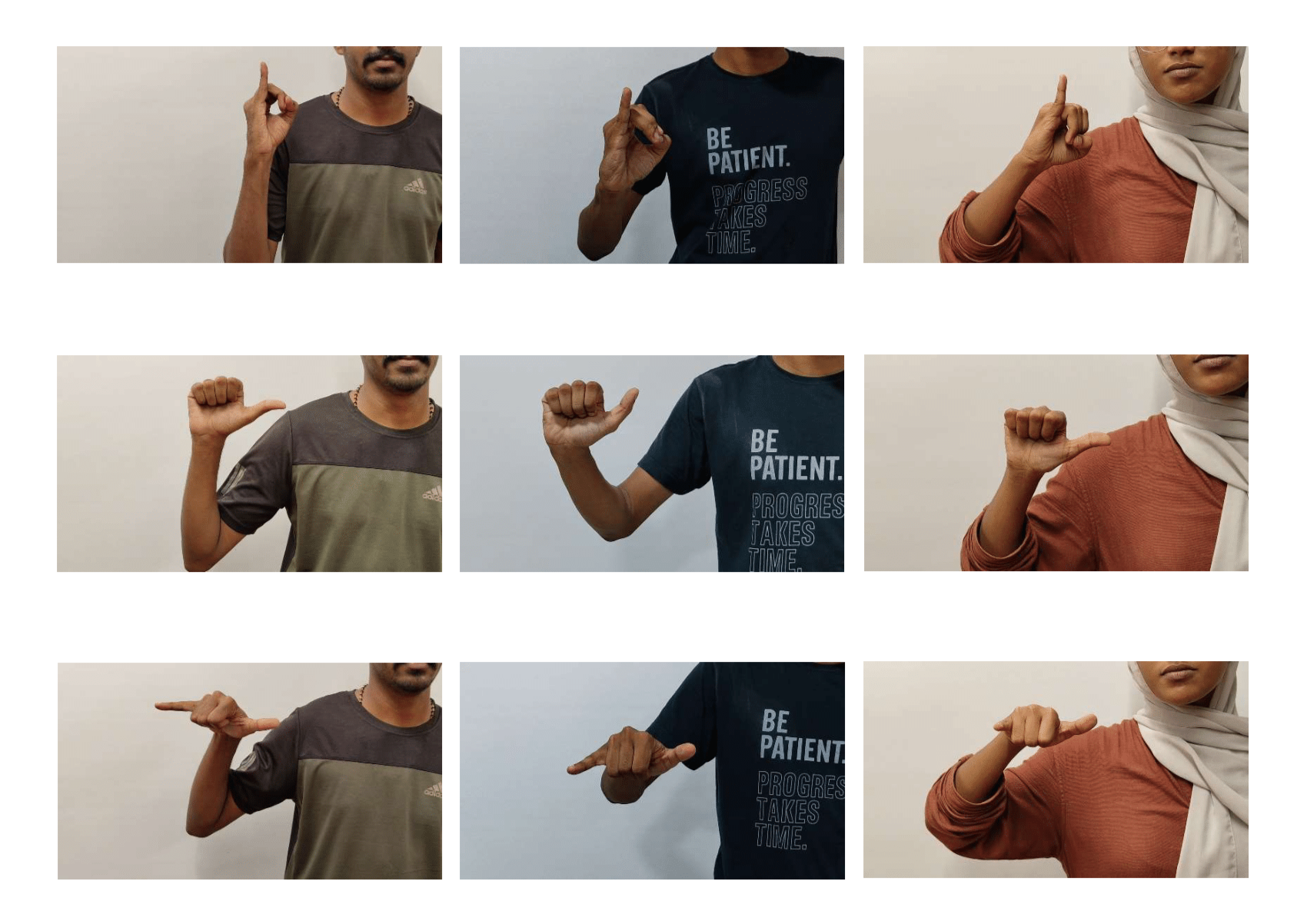}
    \caption{Sample images from the dataset created} 
    \label{fig:samples}   
\end{figure}

\section{Proposed Methodology}
This section details our proposed methodology for the automatic detection of vehicle and identification of number plates using convolutional neural networks and YOLOv8. The overall workflow of the proposed approach is given in Fig. 4.\\

\label{ref_propapp}
\begin{figure}[h]
    \centering
    \includegraphics[width=0.85\linewidth]{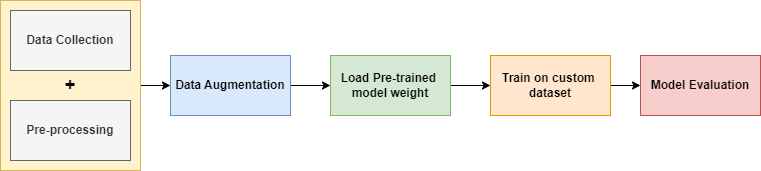}
    \caption{The overall workflow of sign language detection using pre-trained model} 
    \label{fig:flowchart}   
\end{figure}

For maintaining better accuracy, a specific dataset is developed for the project. Twenty signs from 20 different people are collected by recording videos in 60 fps with 4k resolution. Each sign consist of almost 300 frames and the whole dataset comprises of 5900 data points. In the preprocessing stage, the image is resized to 432 x 256 pixels to reduce the size of the dataset, thus helping make the computation much easier. Data augmentation is a technique used in deep learning to improve the quality of the data being used to train a model. It does so by adding small variations to existing data samples. This helps prevent the models from overfitting and improves the model accuracy. We have used two augmentation techniques have been used such as noise - that add random noises to the sign images, and rotation - that rotates the images in a specified angle to the right or left.

By utilizing the pretrained weights from the model YOLOV8, it enables the transfer learning, accelerating model convergence and improving generalization. This initialization aids the YOLOv8 model in grasping common patterns, reducing the need for extensive training iterations. Specifically within the YOLOv8 architecture, loading pretrained weights allows the model to leverage insights from a broader object detection task, thereby improving its proficiency in recognizing and identifying objects in specific images. The pre-initialized weights provide a strong starting point, significantly reducing training time compared to random initialization. The inherited knowledge often leads to higher accuracy in the new model, as it leverages the features and relationships that have already been discovered. The new model inherits the pre-trained model's ability to generalize to new data not encountered during its own training. This leads to improved performance on diverse tasks and situations. Now the augmented dataset labeled on YOLOV8 format is trained with the pre-trained YOLOV8 model. Later initiated the training process adjusts the epochs and evaluates the model performance.

\section{Experiments}
This project was done on a system with an Intel Core i5 processor running Windows 11. The code was executed in Google Colab with GPU support. Initially, we collected nearly 2000 images of 20 different hand gestures from 20 different individuals as described in Section 2. Subsequently, we resized the images to 432x256 pixels. The next step involved data augmentation.
\subsection{Data Augmentation}
The process in which the training dataset is converted to a new synthetic data sample by adding small perturbations is called Data Augmentation. It is done by applying noise injection, and rotation. It is to make the model invariant to those perturbations and enhance its generalization ability. In this dataset, we added 5\% noise and 10 degrees positive and negative rotation to the dataset (Refer Section \ref{ref_DataAugmentation}.)

\subsection{Load and build Ultralytics YOLOv8 model}
Ultralytics YOLOv8 is an open-source deep learning framework designed for object detection and more complex tasks. Built upon the YOLO architecture, YOLOv8 by Ultralytics is known for its efficiency and speed in real-time object detection. It incorporates advancements in model architecture, training strategies, and deployment options, making it a popular choice for computer vision applications. The framework is implemented in PyTorch and offers a user-friendly interface, enabling researchers and developers to easily train and deploy object detection models with high accuracy and performance.

Initially, we loaded a pre-trained YOLOv8 model. Subsequently, we input our custom dataset into the pre-trained model, initiating the training process for a YOLOv8 object detection model. The 'model' object, assumed to be an instance of YOLOv8, is invoked with the 'train' method. The 'data' parameter points to a YAML file, likely containing dataset configuration details such as file paths and class labels. The training is set to run for 100 epochs, with early stopping implemented using a patience value of 10 epochs. The verbose setting indicates detailed information about the training progress, including loss values and metrics. The 'best.pt' folder is automatically generated by YOLO and contains weights stored in the 'runs/train/weights' path. The best weights are saved under the name 'best.pt.' The 'best.pt' checkpoint is selected based on a predefined metric, such as accuracy or loss, achieved on a validation set. By choosing this checkpoint for deployment, the code ensures that the model used in production is the most effective version, as determined by its performance on unseen data.

\section{Results and Discussions}
\label{ref_results}
\begin{figure}[h]
    \centering
    \includegraphics[width=0.7\linewidth]{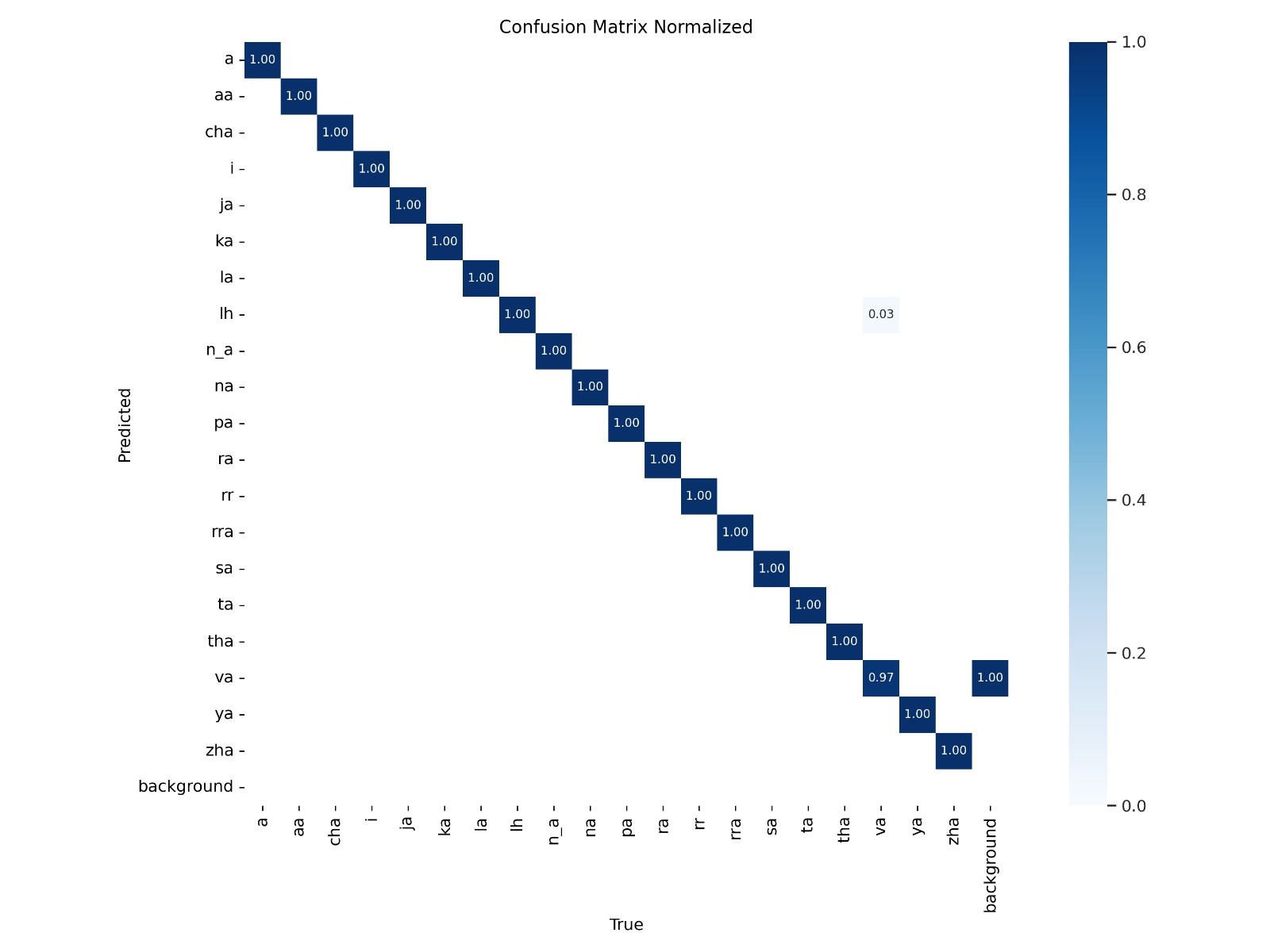}
    \caption{Confusion matrix of the model} 
    \label{fig:confusionmatrix}   
\end{figure}

Fig \ref{fig:confusionmatrix} illustrates the confusion matrix for the classification of 20 classes/signs using YOLOv8 on the
validation data. Predicted classes are on the vertical axis, while ground truth is along horizontal 
axis. It can be observed that most of the classes are correctly classified by YOLOv8 except 'va' which has a value of 0.97. In a heatmap, the diagonal cells from the top-left to bottom-right represents the correct predictions. The darker these cells, the better the model performs. Evidently, all the diagonal cells of the confusion matrix are darker  than rest of the cells, meaning the model is performing well and making predictions accurately. \\

\begin{figure}[h]
    \centering
    \includegraphics[width=0.3\linewidth]{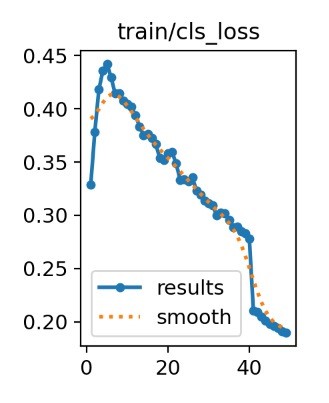}\hfill
    \includegraphics[width=0.3\linewidth]{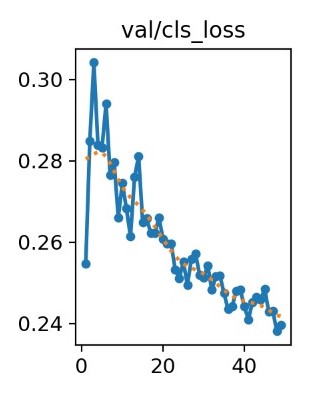}
    \includegraphics[width=0.3\linewidth]{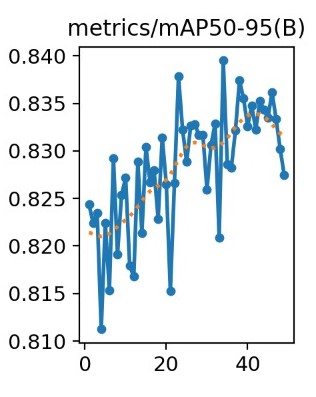}
    \caption{Training and Validation loss plots} 
    \label{fig:samples}   
\end{figure}

Initially, at epoch 1, training loss is comparatively low. This is because the model has not yet learned any patterns in the data. As training progresses (from epochs 2 to 8), the training loss 
increases rapidly and decreases from an epoch 9. This indicates that the model is improving its fit to the training data and is  becoming proficient at recognizing patterns within it. Similarly, the validation loss also starts at lower value during epoch 1, as the model has not been exposed to the validation data. However, as training continues (from epochs 2 to 8), the validation loss initially increases and decreases from there on. This is a positive sign, indicating that the model is generalizing well to unseen data. 
Also, a smooth, downward-trending line indicates that the model is learning and improving. It means the error between its predictions and the ground truth is decreasing over time. Mean Average Precision (mAP) is a metric used to evaluate object detection models. It incorporates the trade-off between precision and recall and considers both false positives and negatives. This property makes mAP a suitable metric for most detection applications. As shown in Figure \ref{fig:acc_plot}, throughout the training progression, there is a discernible and consistent rise in mAP, despite occasional fluctuations. The model achieves an mAP value of approximately 83.9\%.\\

\begin{figure}[h]
    \centering
    \includegraphics[width=0.8\linewidth]{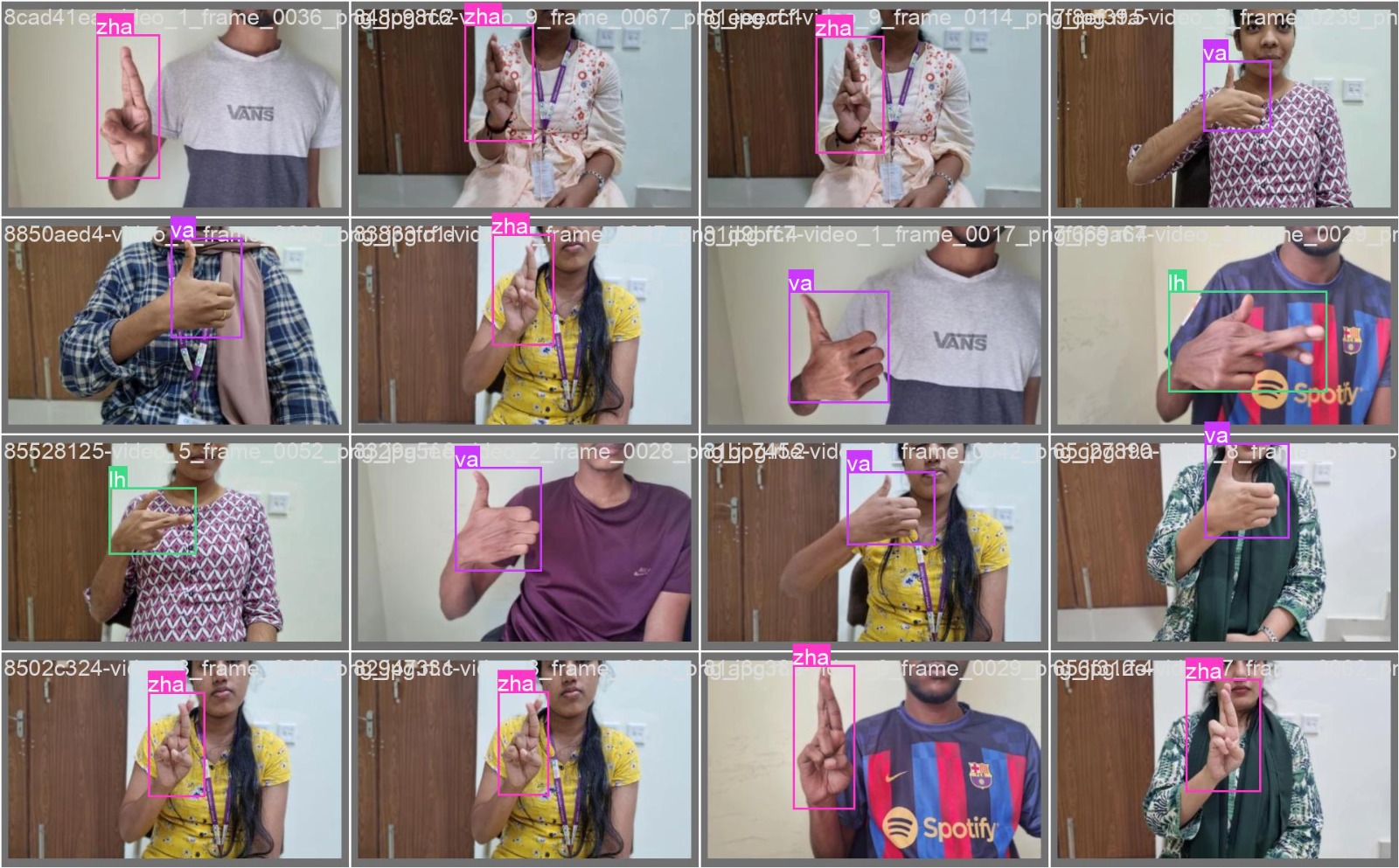}
    \caption{Validation batch (labels)} 
    \label{fig:val_labels}   
\end{figure}
\begin{figure}[htp!]
    \centering
    \includegraphics[width=0.8\linewidth]{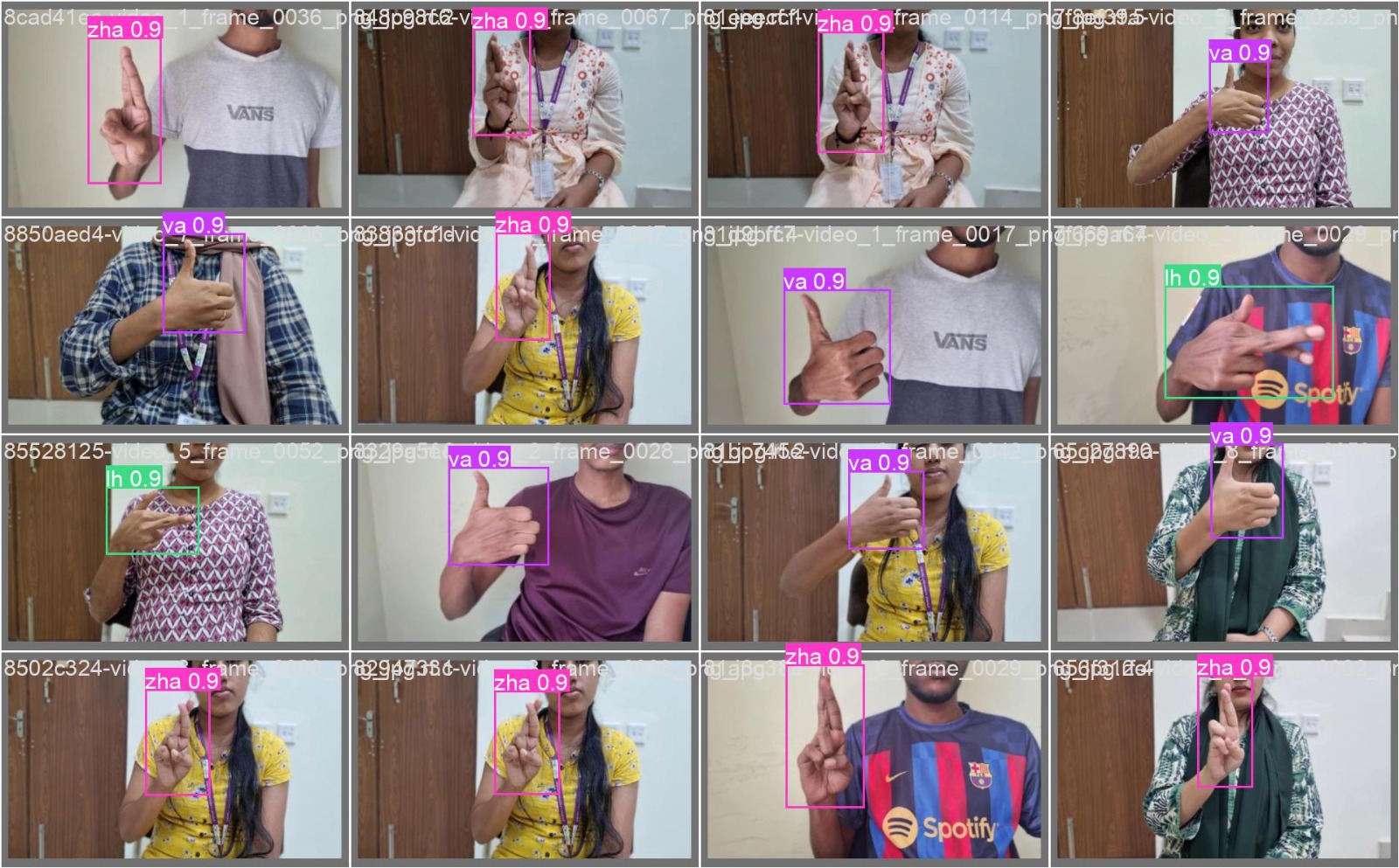}
    \caption{Validation batch (predicted)} 
    \label{fig:val_preds}   
\end{figure}
Fig \ref{fig:val_labels} shows corresponding labels of the sign which is in validation data represented by a bounding box and the \ref{fig:val_preds} illustrates the probability of predicting the sign correctly.Most of the probability lies in the range of 0.8 to 0.9 which shows that the model yields high accuracy.

\section{Conclusions and Future Work}
This research aims to craft an advanced Malayalam Sign Language Identification system, designed to effectively detect static gestures. Beyond the fundamental translation of these gestures into captions, the system operates in real-time, leveraging the power of computer vision and deep learning to achieve a commendable level of accuracy. In terms of future scope, the planned expansion involves enhancing the system's capabilities to encompass dynamic signs. The potential applications of this model are broad and impactful. Beyond its immediate utility, the system could find integration into emergency response systems, offering a vital communication tool in critical situations. Moreover, its implementation in video calls could foster inclusive communication, breaking down barriers for individuals with hearing impairments. The model's adaptability extends to educational settings, contributing to an inclusive learning environment for those utilizing Malayalam sign language. Furthermore, it could play a crucial role in the development of innovative technologies tailored to the unique communication needs of the Malayalam sign language community.
\bibliographystyle{unsrtnat}
\bibliography{references} 
\end{document}